\pdfoutput=1
\documentclass{article} % For LaTeX2e
\usepackage{iclr2027_conference,times}

\usepackage{amsmath,amsfonts,bm}

\def\eqref#1{equation~\ref{#1}}
\def\1{\bm{1}}

\DeclareMathAlphabet{\mathsfit}{\encodingdefault}{\sfdefault}{m}{sl}
\SetMathAlphabet{\mathsfit}{bold}{\encodingdefault}{\sfdefault}{bx}{n}

\DeclareMathOperator*{\argmax}{arg\,max}

\usepackage{hyperref}
\usepackage{url}

\usepackage{graphicx}
\usepackage{caption}
\usepackage{subcaption}

\usepackage{algorithm}
\usepackage{algorithmic}

\usepackage{booktabs}
\usepackage{multirow}

\title{Going Beyond State-Reaching: Learning \\Abstractions for Intrinsically Motivated \\Option Discovery}

\author{Akhil Bagaria\thanks{Equal contribution.}~~\thanks{Work done while at Brown University.} \\
Amazon \\
\texttt{akhilbg@amazon.com} \\
\And
Anita De Mello Koch\footnotemark[1] \\
Brown University \\
\texttt{anita\_de\_mello\_koch@brown.edu} \\
\And
George Konidaris \\
Brown University \\
\texttt{gdk@brown.edu} \\
}

\iclrfinalcopy % arXiv version is non-anonymous; header patched to `Preprint.' in the local .sty.

\begin{document}

\maketitle

\begin{abstract}
Temporal abstraction via options can improve exploration in large environments. However, existing option discovery algorithms find subgoals that target all aspects of the state simultaneously. This \textit{state-reaching} approach produces options that only apply in narrow regions of the state-space, eventually causing an explosion in the number of options that overwhelms the agent, and impedes progress on its primary task of reward maximization. We introduce an algorithm that instead identifies a small, relevant subset of features for each subgoal, yielding options that generalize broadly and accelerate exploration. Our approach learns abstract, transferrable options and achieves rapid exploration in three sparse-reward, image-based domains, including the Atari game \textsc{MontezumasRevenge}.
\end{abstract}

\section{Introduction} \label{introduction}

Reinforcement learning (RL) agents must interact with the environment at every timestep. Yet, effective decision-making may demand temporal abstractions to ease central challenges of credit assignment, exploration, and transfer \citep{HRLSurvey2025}. The core question is---how can RL agents autonomously discover temporally extended actions, or \textit{options} \citep{sutton1999between}, solely via environment interaction? This problem---skill or option discovery---is the core bottleneck for scaling hierarchical RL (HRL; \citealp{barto2003recent}) to long-horizon, high-dimensional problems \citep{konidaris2019necessity,gershman2017blessing,sutton2022alberta}. However, most skill discovery algorithms today have a common shortcoming: they resort to the state-reaching objective, i.e., they create skills that target entire states, rather than small, relevant portions of those states. 

% Placed in-line (not floated) so that Figure 1 is guaranteed to appear on page 1.
\begin{center}
    \includegraphics[width=\linewidth]{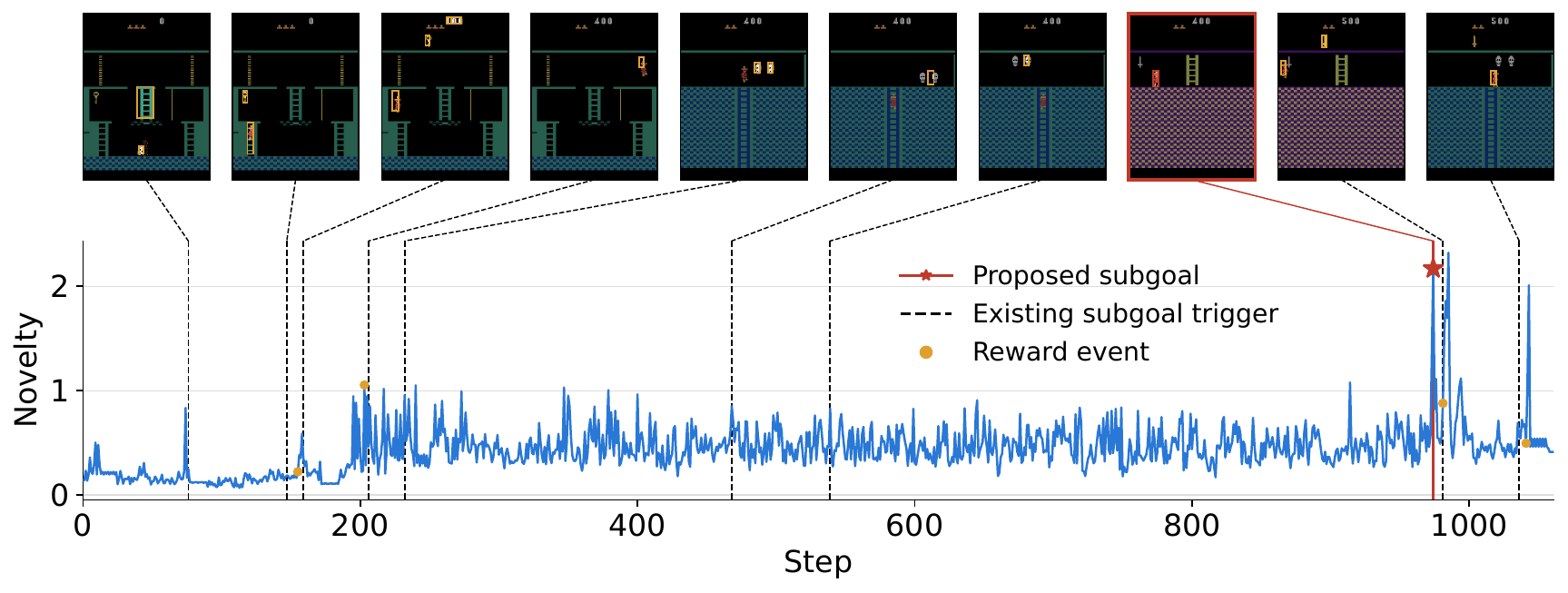}
    \captionof{figure}{\textbf{Abstract subgoal discovery in a \textsc{MontezumasRevenge} episode.} The blue curve shows novelty over an episode. Frames show achieved option subgoals; yellow bounding boxes isolate the parts of the image the option's subgoal classifier attends to, ignoring all other pixels. The frame with the red box denotes the subgoal for a new option discovered in this episode.}
\end{center}

To understand state-reaching, recall the moment you first balanced a bicycle without training wheels and realized you had done something right. Practicing this new skill later, you recreated your speed and balance, but not the color of your t-shirt, the weather, or the exact wear of your tires. By contrast, existing skill discovery algorithms that learn to recreate a previously encountered experience \citep{chentanez2005intrinsically} attempt to recreate \textit{every aspect} of that state. Examples include the $\epsilon$-ball from deep skill graphs \citep{bagaria2021dsg}, pixel-wise equality \citep{veeriah2018many}, the perceptual hashing from Go-Explore \citep{ecoffet2021first}, random projections of the target state \citep{dabney2021vip,farebrother2023protovalue}, or a neural classifier that predicts perceptual similarity between states \citep{vezhnevets2017feudal,hafner22director}. This state-reaching strategy hurts hierarchical agents in several ways. First, the resulting skills are nontransferable; consider a robot grabbing a cup from a cluttered table: the skill's subgoal is tied to the position of every object in the state, requiring the agent to learn combinatorially more skills. Beyond stymieing transfer---a key benefit of HRL \citep{taylor2009transfer}---state-reaching artificially reduces the size of the skill's subgoal region, complicating policy learning and eventually making the skill less useful. Finally, recreating a state is \textit{unscalable} because in large, realistic domains, recreating all aspects of a particular state may simply be impossible \citep{bagaria23scaling,javed2024the}.

Our skill discovery algorithm is inspired by intrinsic motivation \citep{oudeyer2008can,colas2022autotelic}: during play, when children cause something interesting to happen, they try to recreate it, with increasing efficiency \citep{white1959motivation}. They practice until they get bored, then move on, retaining the new skill for later reuse \citep{barto2004intrinsically}. Similarly, when an AI agent's exploration reveals something particularly interesting, it should learn a skill to reliably achieve it again. Other algorithms (e.g., \citealp{bagaria2021dsg,bagaria2025intrinsically,Jinnai2020Exploration}) use these ideas, but resort to a state-reaching objective. Instead, when the agent encounters a state that produces a spike in novelty \citep{chentanez2005intrinsically}, we use \textit{feature attribution} to identify the features responsible, and construct a subgoal classifier that attends only to those specific features.\footnote{We do not assume that state features are given---instead learning subgoal achievement classifiers directly from images---but the language of ``features'' helps present our ideas simply.} These classifiers serve as reward signals for training each option's policy, and a high-level policy selects among these resulting options to maximize the agent's overall reward.

We test our algorithm in three challenging exploration problems: \textsc{MiniGrid-KeyCorridor}, \textsc{VisualTaxi}, and the Atari game \textsc{MontezumasRevenge}---all with image-based observations and sparse reward functions. We focus on image-based domains to stress-test our algorithm: when a factored state representation is available, each dimension serves as a candidate feature and a streamlined version of our algorithm applies directly. Our method outperforms both non-hierarchical (flat) RL baselines and a state-reaching hierarchical baseline that shares our subgoal discovery algorithm but builds classifiers over the entire image. Our discovered subgoals depend on only a small number of features and are achieved across qualitatively different states, enabling skill reuse.

\paragraph{Contributions.} Conceptually, we identify the implicit state-reaching assumption shared by existing option discovery methods and argue that identifying a relevant subset of features per subgoal is a crucial missing step. Methodologically, we introduce an algorithm that detects novelty spikes during exploration, attributes them to a small set of features, and constructs subgoal classifiers that attend only to those features, resulting in options that can be reused in different contexts. Empirically, we show on \textsc{MiniGrid-KeyCorridor}, \textsc{VisualTaxi}, and \textsc{MontezumasRevenge} that this abstraction outperforms flat-RL baselines and a state-reaching hierarchical baseline.

\section{Background and Related Work} \label{sec:background}
We model the agent-environment interaction as a Markov Decision Process ${M=(S, A, R, T, \gamma)}$ \citep{Puterman94mdp,sutton2018reinforcement}. To aid in reward maximization, our agent will discover options $o = (\mathcal{I}_o, \pi_o, \beta_o) \in \mathcal{O}$ \citep{sutton1999between}, where $\mathcal{I}_o: S \rightarrow [0, 1]$ is the initiation function, $\pi_o: S \rightarrow A$ is the intra-option policy over primitive actions, and $\beta_o: S \rightarrow \{0, 1\}$ is the termination condition. Each option has a maximum horizon of $H$ steps.

\paragraph{Subgoal Options.} While options need not optimize any objective in general, for option \textit{discovery} it is convenient to think of options as achieving subgoals---indeed, most option discovery algorithms can be viewed this way \citep{precup2001temporal,bagaria2024skill,HRLSurvey2025}. A subgoal here is not a state, but is instead a classifier that grounds to a potentially large set of states. A \textit{subgoal option} uses this classifier to serve three roles: it defines the option's subgoal, its termination condition $\beta_o: S \rightarrow \{0,1\}$, and provides a binary pseudo-reward $R_o(s)=\beta_o(s)$ for training $\pi_o$. The initiation function $\mathcal{I}_o$ identifies states from which the option has a high probability of achieving its subgoal. This formulation reduces option discovery to identifying useful subgoals: once $\beta_o$ is specified, standard policy optimization can train $\pi_o$, and policy evaluation can train $\mathcal{I}_o$.

\paragraph{Universal Value Function Approximators (UVFAs)} \citep{schaul2015universal} are a scalable way to learn goal-conditioned value functions $V_g: S\rightarrow \mathbb{R},\; Q_g: S\times A \rightarrow \mathbb{R}$ and their corresponding goal-conditioned policies $\pi_g: S \rightarrow A$ \citep{kaelbling1993learning} using function approximation. We use UVFAs to parameterize option value functions and policies: instead of representing each option's policy with a separate function approximator \citep{sutton2011horde}, a single function approximator can be shared across all options by conditioning it on each option's own subgoal \citep{bagaria2021robust}: ${\pi_o(s) = \argmax_a Q_{g}(s, a) = \argmax_a Q_{\beta_o}(s, a)}$.

\paragraph{Novelty-based exploration.} In tabular RL, count-based exploration bonuses that decay as $1/\sqrt{N[s, a]}$ can guide efficient learning \citep{strehl2008analysis}, but do not scale to large state spaces. Pseudocounts \citep{bellemare2016unifying} generalize counts by assigning similar values to similar states; Coin Flip Networks (CFN; \citealp{lobel2023flipping}) are a simple, state-of-the-art technique for estimating pseudocounts from images. Like other pseudocount methods, CFN learns a function $f_\phi^{\text{int}}$ that maps states to an approximate $1/\sqrt{N}$ bonus. We use CFN over alternatives like Random Network Distillation (RND; \citealp{burda2018exploration}) because \citet{lobel2023flipping} showed it outperforms popular novelty estimators---especially in stochastic domains---on task reward, hyperparameter robustness, and exploration-bonus interpretability. Our framework, however, is agnostic to the choice of novelty estimator. Importantly, while novelty-driven agents demonstrate impressive performance \citep{kapturowski2022human}, they do not learn options, which are key for transfer \citep{taylor2009transfer} and high-level planning \citep{konidaris2018skills,sutton2024reward}. We seek to bridge this gap by using novelty as the substrate for option discovery~\citep{simsek04relative}. 

\paragraph{Gradient-based feature attribution.} When multiple features jointly produce an outcome, how should we attribute credit to each one? Gradient-based attribution methods~\citep{ancona2017towards} estimate each feature's contribution directly from a model's gradients. One possible target for such methods is the \textit{Shapley value} \citep{shapley1953value} from cooperative game theory, which measures each feature's \textit{marginal contribution}---how much the outcome changes when that feature is added---averaged over all possible orderings of features. Concretely, for some set function $v$ over features $\{1, \ldots, k\}$, the Shapley value of feature $i$ is:
\begin{equation}
    \psi_i = \frac{1}{k!} \sum_{\rho} \Big[ v\big(B_i^\rho \cup \{i\}\big) - v\big(B_i^\rho\big) \Big],
\end{equation}
where $\rho$ ranges over permutations of features and $B_i^\rho$ is the set preceding $i$ in $\rho$. Computing exact Shapley values requires evaluating $v$ on exponentially many subsets, which is intractable for high-dimensional inputs such as images.
So, we use the DeepSHAP algorithm \citep{Lundberg2017AUA,shrikumar2017learning}, which efficiently approximates Shapley values for neural networks by backpropagating attributions from the output to per-pixel contributions.

\paragraph{Option discovery methods.} Option discovery techniques can be broadly categorized by their learning signal. Option-Critic \citep{bacon2017option} and feudal approaches \citep{dayan1993feudal,vezhnevets2017feudal} learn options using extrinsic reward, but struggle in sparse-reward settings. Empowerment-based methods \citep{eysenbach2018diversity,gregor2016variational} learn diverse skills that increase the agent's control over its environment, showing promise in exploration \citep{hansen2021entropic,campos2020explore}. Spectral methods use the graph Laplacian to find principal directions of the state-space for exploration \citep{machado2017Laplacian,jinnai2019covertime,klissarov2023deep}. Some other algorithms also combine goal-conditioned RL (GCRL) and novelty \citep{pitis2020mega,pong2019skew,simsek04relative,bagaria2025intrinsically}. While these approaches differ in their objectives, they share a common limitation: most skills resort to state-reaching because they attend to all features. Our work is orthogonal---we focus on \textit{how} subgoals are represented, showing that identifying relevant features enables broader generalization. For a detailed survey of option discovery, we refer readers to \citet{HRLSurvey2025}.

\paragraph{Learning Subgoal Achievement Classifiers.} A subgoal achievement classifier can either be hand-engineered or learned. Hand-engineered classifiers rely on privileged information: Go-Explore~\citep{ecoffet2021first} uses hand-crafted cell representations, and many GCRL methods~\citep[e.g.,][]{eysenbach2018diversity,nachum2018data,andrychowicz2017hindsight} define subgoals over a designer-chosen subset of variables---like the $(x, y)$ location of the agent in navigation tasks. This sidesteps, rather than solves, the problem of deciding which features a subgoal should attend to. Learned classifiers require no such privileged information, but typically attend to \emph{all} state features: perceptual similarity approaches threshold distances in observation space using random projections \citep{dabney2021vip} or autoencoders \citep{tang2017exploration}; temporal distance methods predict whether states are within a fixed number of steps of a goal \citep{savinov2018semiparametric,mendonca21discovering}; and mutual information methods like DISCERN \citep{wardefarley2019discern} abstract away uncontrollable features but not irrelevant, controllable ones. Attending to every feature is state-reaching, and the benefits of abstraction fade \citep{colas2022autotelic}. In contrast, our method \emph{learns} which small set of features matters for each subgoal.

\paragraph{Options for feature attainment.} The strategy of learning options that target specific features has been explored in HRL \citep{hengst2002hexq}. STOMP \citep{sutton2024reward} learns one option per feature, each option maximizing its feature. Proto-goal RL \citep{bagaria23scaling} generalizes this to real-valued targets over combinations of features. Both assume a factored state representation is provided and treat every feature, or every combination, as a subgoal. Our approach starts from candidate features extracted from pixels and selects, for each subgoal, the few that account for a novelty spike. Our subgoal classifiers are also predicates in the sense of symbolic planning---boolean functions of a few state variables \citep{fikes1971strips,ghallab2004automated}. In that field, either the predicates~\citep{yang2018peorl,lyu2019sdrl}, or skills~\citep{konidaris2018skills,silver2023predicate,alper2025neuro}, or both~\citep{kaelbling2011hierarchical,garrett2021integrated} are given. Our agent learns both via online exploration, extending this bridge between online RL and symbolic planning.

\section{Method: Abstract Subgoal Discovery}

Modern RL systems can efficiently compute proxies for intrinsic motivation, like novelty \citep{kapturowski2022human}. However, when used for option discovery, these signals typically drive state-reaching: the agent is rewarded for re-visiting states that previously yielded high novelty or reward, which fails to scale beyond modest navigational problems \citep{hansen2021entropic}. Instead, our agent measures the \textit{interestingness} of a state using a learned intrinsic reward model $f_\phi^{\text{int}}$ that estimates novelty (Section~\ref{sec:background}). When the agent encounters a state $s_N$ in which novelty spikes substantially relative to the least interesting states $\mathcal{S}_B$ from the same trajectory---states where nothing \textit{interesting} or novel was happening--- it asks: which features of $s_N$ account for the expected spike $\Delta_n=\mathbb{E}_{s_B\sim\mathcal{S}_B}[f_\phi^{\text{int}}(s_N)-f_\phi^{\text{int}}(s_B)]$? Features whose modification substantially changes $\Delta_n$ are retained; the rest are discarded. A classifier is then constructed to attend only to these relevant features, serving as an abstract subgoal (illustration in Figure~\ref{fig:factored-clfs/illustration}, Appendix~\ref{sec:core-insight}). The agent learns an option policy to achieve this subgoal, and the cycle repeats: the agent explores from promising subgoals, discovers new interesting states, and creates new options.

\subsection{Option Selection and Exploration}\label{sec:exploration}

Our agent maintains a set of options $\mathcal{O}$; each option in that set has a subgoal classifier $\beta_o$ discovered during exploration. Every episode, the agent selects a subgoal to explore from, navigates there using the corresponding option policy, and then executes a novelty-driven exploration policy $\pi_{\text{int}}$ \citep{ecoffet2021first,mendonca21discovering,bagaria2025intrinsically}.

\paragraph{Policy over options.} Not all subgoals are equally useful to pursue: some are unreachable from the current state, others lead to already-explored regions. The policy over options $\pi_\mathcal{O}(o \mid s)$ balances achievability, task performance, and exploration potential. Given that $V_{\beta_o}(s)$ can be interpreted as a discounted probability of reaching the option's subgoal $\beta_o$ from state $s$ \citep{bagaria2023effectively}, the agent restricts option selection using the following initiation function:
\begin{equation}\label{eq:initiation-set}
    \mathcal{I}_o(s) = \mathbb{I}\{V_{\beta_o}(s) > \delta\},
\end{equation}
where $\delta \in \mathbb{R}$ is a threshold hyperparameter. Among options for which $\mathcal{I}_o(s) = 1$, the agent scores each by balancing two terms: (1) the expected extrinsic reward $\hat{R}_o$ accumulated during option execution, and (2) the value of exploring from the subgoal:
\begin{align}\label{eq:bandit-utility}
    U(o) &= \alpha\hat{R}_o + \mathbb{E}_{s' \,:\, \beta_o(s')=1} \big[V_{\text{int}}(s')\big],\\
    \text{where } V_{\text{int}}(s') &= \mathbb{E}_{\pi_{\text{int}}}\left[\sum_t \gamma^t \big(r_t + \lambda f_\phi^{\text{int}}(s_t)\big)\right],
\end{align}
and $\alpha, \lambda \geq 0$ control the tradeoff between extrinsic and intrinsic rewards; we used $\lambda=0.01$ in all our experiments. We implement the option reward model $\hat{R}_o$ using a simple average of extrinsic returns over past option executions. The policy is a bandit that samples proportionally to utility:
\begin{equation}\label{eq:policy_over_options}
    \pi_\mathcal{O}(o \mid s) = \frac{U(o)}{\sum_{o' : \mathcal{I}_{o'}(s) = 1} U(o')}.
\end{equation}

\paragraph{Conducting exploration.} Once an option $o$ is sampled from $\pi_\mathcal{O}$, the agent executes its intra-option policy $\pi_{o}$ to reach the subgoal. Upon success, the agent executes the low-level exploration policy $\pi_{\text{int}}$ that maximizes intrinsic and extrinsic reward, producing trajectory $\tau = (s_1, \ldots, s_K)$.

\paragraph{Triggering Option Discovery.} After each exploration rollout, the agent checks whether the trajectory $\tau$ contains a sufficiently large spike in novelty. Concretely, it tests whether $\max_{s \in \tau} f_\phi^{\text{int}}(s) > \mu + \sigma_{\text{state}}\,\sigma$, where $\mu$ and $\sigma$ are the running mean and standard deviation of $f_\phi^{\text{int}}$, and $\sigma_{\text{state}}$ is a threshold multiplier ($1$ in all our experiments; ablated in Appendix~\ref{app:ablations}). If the test passes, the agent extracts the most novel state, $s_N = \argmax_{s \in \tau} f_\phi^{\text{int}}(s)$, encountered at timestep $t_N$. It then constructs a set of baseline states $\mathcal{S}_B=\{s_B^1,\ldots,s_{B}^M\}$. Each baseline is drawn (without replacement) from a window of $w$ timesteps around $t_N$: the agent repeatedly selects the lowest-novelty state whose novelty falls at least $\sigma$ below $f_\phi^{\text{int}}(s_N)$. Finally, $s_N$ and $\mathcal{S}_B$ are passed to the feature extraction algorithm (Section~\ref{sec:feature-selection}).

\subsection{Identifying Relevant Features}\label{sec:feature-selection}

Given $s_N$ and $\mathcal{S}_B$ from the previous section, the agent identifies which features of $s_N$ explain the spike $\Delta_n$ defined as follows:
\begin{equation}\label{eq:Delta}
\Delta_n=\mathbb{E}_{s_B\sim\mathcal{S}_B}[f_\phi^{\text{int}}(s_N)-f_\phi^{\text{int}}(s_B)]\approx f_\phi^{\text{int}}(s_N)-\frac{1}{M}\sum_{j=1}^Mf_\phi^{\text{int}}(s^{j}_B).
\end{equation}
This is further decomposed into two steps: extracting candidate features from the image observation, and then measuring each feature's contribution to $\Delta_n$.

\paragraph{Feature extraction.} Neural representations learned end-to-end are entangled, making it difficult to intervene on individual factors \citep{RodriguezSanchez2025FromPT,Kumar2025QuestioningRO}. Our method relies on disentangled candidate features, and can leverage any extraction technique capable of producing them. We explore two off-the-shelf computer vision techniques: a lightweight contour detection algorithm~\citep{suzuki1985topological}, and a pretrained segmentation model~\citep{kirillov2023segany}. The output of this module is a set of candidate features $\{b_1,...,b_k\}$ to be used for attribution.  

\begin{figure}[h!]
    \centering
    \includegraphics[width=.85\linewidth]{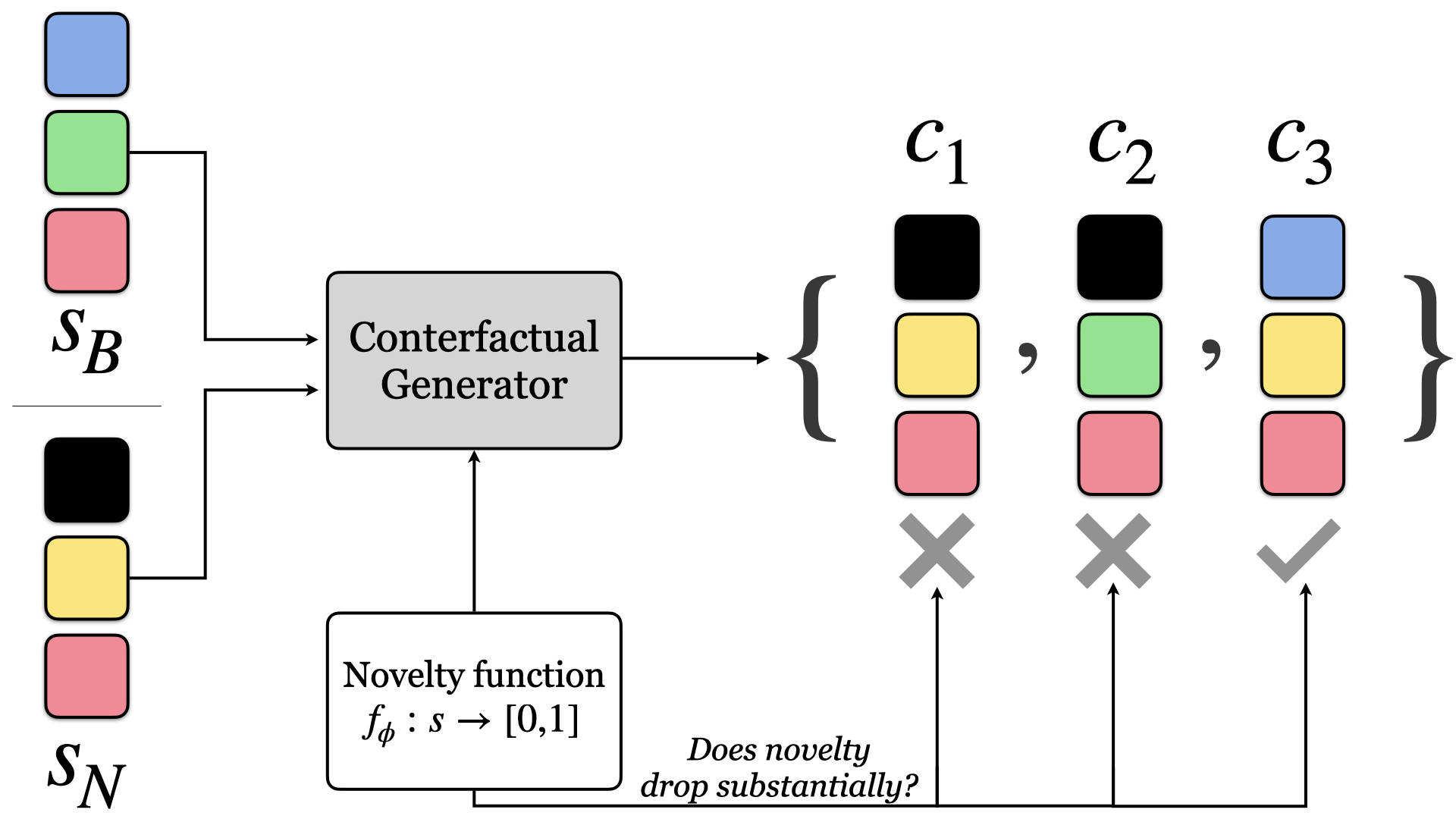}
    \caption[Subgoal classifier feature selection]{\textbf{Illustration of counterfactual substitution.} The most boring and most novel states, $s_B$ and $s_N$ respectively, are input into the Counterfactual Generator, which outputs counterfactual states $c_1, c_2, c_3$—each with one feature reset to its value in $s_B$. Only resetting the first feature (black $\rightarrow$ blue) substantially lowers novelty, indicating that the first feature accounts for most of the change in novelty $\Delta_n$. This one-at-a-time reset is effective when features contribute independently.}
    \label{fig:factored-feature-selection}
\end{figure}

\paragraph{Feature attribution.} Given candidate features $\{b_1,...,b_k\}$, the agent must determine which ones explain the novelty difference. The core question is: how much does each feature $b_i$ contribute to $\Delta_n$? We consider two techniques. \textit{Counterfactual substitution} (see Figure \ref{fig:factored-feature-selection}) uses a single baseline $\mathcal{S}_B=\{s_B\}$ and constructs a counterfactual state $c_i$ by replacing feature $b_i$ in $s_N$ with the corresponding feature from $s_B$. If the resulting novelty drop $\Delta_i=f_\phi^{\text{int}}(s_N)-f_\phi^{\text{int}}(c_i)$ exceeds a threshold $\epsilon$, then $b_i$ is relevant. This is effective when features contribute to $\Delta_n$ roughly independently.
\textit{Gradient-based attribution methods} \citep{ancona2017towards} estimate each feature's contribution directly from $f_\phi^{\text{int}}$'s gradients, without requiring explicit counterfactual generation. These methods return an attribution, $a_p$, for each pixel $p$. We convert this to a feature-level attribution by averaging over the pixels in $b_i$: $A_i=\frac{1}{|b_i|}\sum_{p \in b_i}a_p$. A feature $b_i$ is relevant if $A_i>\mu_{A}+\sigma_{\text{patch}}\,\sigma_{A}$, where $\mu_{A}$ and $\sigma_{A}$ are the mean and standard deviation of $\{A_1,\ldots,A_k\}$ over all candidate features and $\sigma_{\text{patch}}$ is a threshold multiplier ($1$ in all our experiments; ablated in Appendix~\ref{app:ablations}). Many algorithms can do this; we use DeepSHAP \citep{Lundberg2017AUA}, with implementation details in Section~\ref{sec:implement}.

This yields a subset of features $\mathcal{F} \subseteq \{b_1, \ldots, b_k\}$ that explain the jump in novelty from $\mathcal{S}_B$ to $s_N$. These features define the scope of the subgoal classifier constructed in the next section.

\subsection{Subgoal Classifier Construction}\label{sec:classifier}

Most of $s_N$ is irrelevant to why it was novel. The feature selection step identified which parts matter; everything else can vary freely. The subgoal classifier respects this---it checks only whether the relevant features $\mathcal{F}$ match their values in $s_N$, and ignores the rest.

The agent constructs a non-parametric classifier $g: S\rightarrow\{0, 1\}$ since $s_N$ is the only positive example. The classifier stores $s_N$ and a feature extractor $h$ that returns only the relevant features---selecting relevant indices for factored states, or cropping bounding boxes for images. For state $s$:
\begin{equation}
    g(s) = \mathbb{I}\Big\{\mathcal{D}\big(h(s_N), h(s)\big) < \xi\Big\},
\end{equation}
where $\xi$ is a threshold hyperparameter and $\mathcal{D}$ measures dissimilarity (e.g., Euclidean distance for factored states, or bounding-box similarity for images). The single-positive-example setting is inherent to the problem---when novelty first spikes, $s_N$ is all that is available---and the feature selection step is precisely what enables the classifier to generalize from it (Section~\ref{sec:experiments}). The classifier $g$ serves as the new option $o$'s termination condition $\beta_o$. This new option $o$ is added to $\mathcal{O}$, making it available for future use by the policy over options $\pi_\mathcal{O}$ (Section~\ref{sec:exploration}).

\subsection{Putting It Together}\label{sec:algorithm}

Algorithm~\ref{alg:abstract-subgoal-discovery} summarizes the full option discovery loop. The agent alternates between exploiting existing options to reach promising subgoals and exploring from there to discover new ones. When exploration yields a state with sufficient spike in novelty, the agent identifies the relevant features, constructs a new subgoal classifier, and adds the corresponding option to its repertoire. 

\begin{algorithm}[h]
\caption{Abstract Subgoal Option Discovery}
\label{alg:abstract-subgoal-discovery}
\begin{algorithmic}[1]
\REQUIRE Algorithm $\mathcal{A}_1$ to estimate novelty $f_\phi^{\text{int}}: S\rightarrow\mathbb{R}^+$.\\
\quad\quad\; Algorithm $\mathcal{A}_2$ to learn intrinsic policy $\pi_{\text{int}}$ and value function $V_{\text{int}}$.\\
\quad\quad\; Algorithm $\mathcal{A}_3$ to learn intra-option policy $\pi_{o}$ and its value function $V_{\beta_o}$.\\
\quad\quad\; Algorithm $\mathcal{A}_4$ to extract features from an image (e.g., contour detection).\\
\quad\quad\; Algorithm $\mathcal{A}_5$ to attribute credit to features (e.g., Shapley values).
\STATE \textbf{Initialize:}
\STATE Initialize option set $\mathcal{O}$ with a default exploration option.
\WHILE{True}
    \STATE Sample option $o \sim \pi_\mathcal{O}(o \mid s)$ using Equation~\ref{eq:policy_over_options}.
    \STATE Execute $\pi_{o}$ until $\beta_{o}(s) = 1$ or timeout after $H$ steps.
    \IF{timeout}
        \STATE \textbf{continue} \COMMENT{Resample option}
    \ENDIF
    \IF{agent reached subgoal}
        \STATE Roll out $\pi_{\text{int}}$ to get trajectory $\tau$.
    \ENDIF
    \IF{$\max_{s \in \tau} f_\phi^{\text{int}}(s) > \mu + \sigma_{\text{state}}\,\sigma$}
        \STATE Extract $s_N = \argmax_{s \in \tau} f_\phi^{\text{int}}(s)$.
        \STATE Construct baseline set $\mathcal{S}_B=\{s_B^1,\ldots,s_{B}^M\}$ (Section~\ref{sec:exploration}).
        \STATE Extract candidate features $\{b_1, \ldots, b_k\}$ from $s_N$ using algorithm $\mathcal{A}_4$.
        \STATE Select relevant features $\mathcal{F}$ using algorithm $\mathcal{A}_5$.
        \STATE Construct classifier $g$ from $\mathcal{F}$ (Section~\ref{sec:classifier}).
        \STATE Add new option $o' = (\mathcal{I}_{o'}, \pi_{o'}, \beta_{o'} = g)$ to $\mathcal{O}$.
    \ENDIF
    \STATE Update mean $\mu$ and standard deviation $\sigma$ of intrinsic rewards.
    \STATE Update novelty estimator $f_\phi^{\text{int}}$ using algorithm $\mathcal{A}_1$.
    \STATE Update $V_{\beta_o}$ and $\pi_o$ using reward $R_o(s)=\beta_o(s)$ and algorithm $\mathcal{A}_3$.
    \STATE Update $V_{\text{int}}$ and $\pi_{\text{int}}$ using reward $r_t + \lambda f_\phi^{\text{int}}(s_t)$ and algorithm $\mathcal{A}_2$.
\ENDWHILE
\end{algorithmic}
\end{algorithm}

\section{Experiments}\label{sec:experiments}

Our experiments address three questions: (1) Does our feature attribution method correctly identify the features responsible for novelty? (2) Do the resulting subgoal classifiers generalize across states that differ in irrelevant features? (3) Does this abstraction lead to better exploration and task performance compared to flat RL and state-reaching HRL?

\subsection{Experimental Design}\label{sec:setup}

\paragraph{Domains.} We evaluate on three sparse-reward, image-based domains where the agent must explore effectively. \textsc{MiniGrid-KeyCorridor-S5R3} \citep{MinigridMiniworld23} requires the agent to navigate through rooms, pick up a key, and unlock a door. \citet{colas2022autotelic} identified it as a challenging exploration problem due to its sparse rewards and combinatorial state space. \textsc{VisualTaxi} \citep{dietterich2000hierarchical} is a $10\times10$ grid-world, in which the agent must navigate to a passenger, pick them up, and deliver them to one of $8$ possible destinations. Variants of this problem have been widely used to evaluate HRL algorithms \citep{dietterich2000hierarchical,allen2021learning}. In both domains, non-zero reward is only observed upon completing the entire task, with no intermediate rewards for picking up the key or passenger. Random exploration is unlikely to ever see non-zero rewards in these domains, making them good testbeds for directed exploration strategies, like ours. \textsc{MontezumasRevenge} \citep{bellemare2013arcade} is a canonical exploration benchmark with a natural hierarchical structure, and we seek to build option discovery algorithms that can autonomously find this hierarchical structure from experience.

\subsection{Implementation details}\label{sec:implement}

\paragraph{Learning policies and value functions.} The classifier $g$ constructed in Section~\ref{sec:classifier} serves as the option's termination condition $\beta_o$ and provides a binary terminating pseudo-reward $R_o(s)=g(s)$. We train the intra-option policy $\pi_o$ using goal-conditioned R2D2 \citep{kapturowski2018recurrent} with hindsight experience replay (HER; \citealp{andrychowicz2017hindsight}), a well-established combination for sparse-reward goal-reaching. Architecturally, we augment the R2D2 CNN encoder with a goal-processing torso; state and goal embeddings are concatenated and passed through an MLP that produces Q-values over actions. The subgoal input to the CNN encoder includes only $s_N$'s extracted features, and $0$ elsewhere. Every episode, the agent uses hindsight relabeling with five subgoals achieved during the trajectory and adds the corresponding sub-trajectories to the replay buffer \citep{andrychowicz2017hindsight}. The initiation function $\mathcal{I}_o(s)$ is derived from the option value function $V_{\beta_o}$, as described in Section~\ref{sec:exploration}. We use CFN~\citep{lobel2023flipping} for learning $f_\phi^\text{int}$ over similar methods~\citep{burda2018largescale} because it provides (1) a more stable range of intrinsic rewards, which stabilizes option learning, and (2) more concentrated Shapley attributions (Appendix~\ref{sec:appendix_cfn_vs_rnd}).

\paragraph{Feature generation and selection.} \textsc{Minigrid} and \textsc{VisualTaxi} have simple image observations with static backgrounds, so we use contour detection~\citep{suzuki1985topological} for feature extraction and counterfactual substitution for feature attribution. This provides a low memory and computation version of the algorithm. 
\textsc{MontezumasRevenge} is visually more complex---contour detection's static background assumption and counterfactual substitution's independence assumption both break---so, we use a pre-trained Segment Anything Model (SAM, \citealp{kirillov2023segany}) for feature extraction and DeepSHAP for feature attribution \citep{Lundberg2017AUA,shrikumar2017learning}. All three domains use the same online training procedure described in Algorithm~\ref{alg:abstract-subgoal-discovery}, with subgoal discovery and option learning occurring jointly. We describe additional \textsc{MontezumasRevenge}-specific implementation details in Appendix~\ref{app:monte}.

\subsection{Qualitative Results}

% We first examine whether our method correctly identifies relevant features and whether the resulting classifiers generalize across visually distinct states.

\begin{figure*}[h]
\centering
\footnotesize
\begin{subfigure}{0.49\textwidth}
    \centering
    \includegraphics[width=\linewidth]{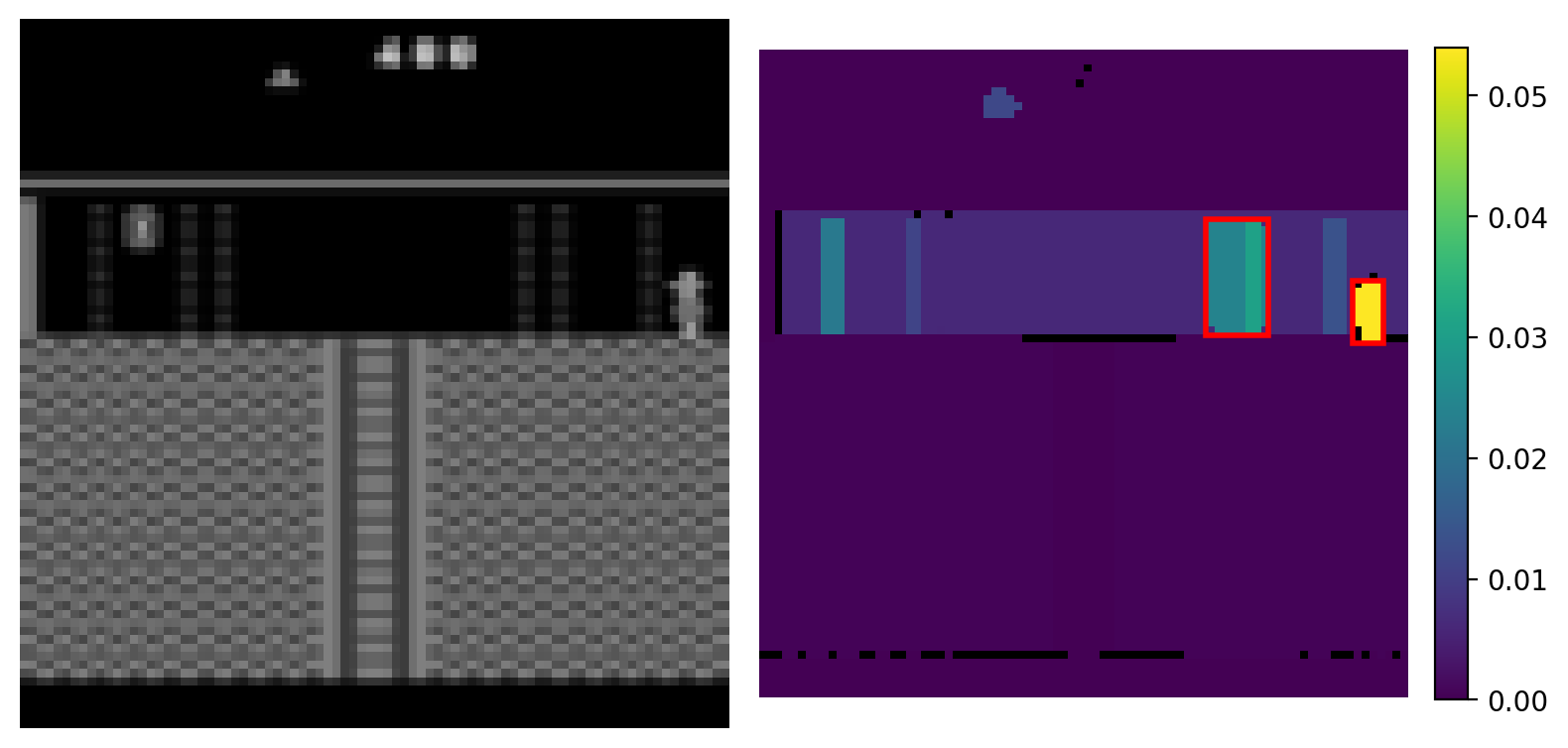}
    \caption{Agent enters room 0.}
    \label{fig:shapley_plots_int}
\end{subfigure}
\begin{subfigure}{0.49\textwidth}
    \centering
    \includegraphics[width=\linewidth]{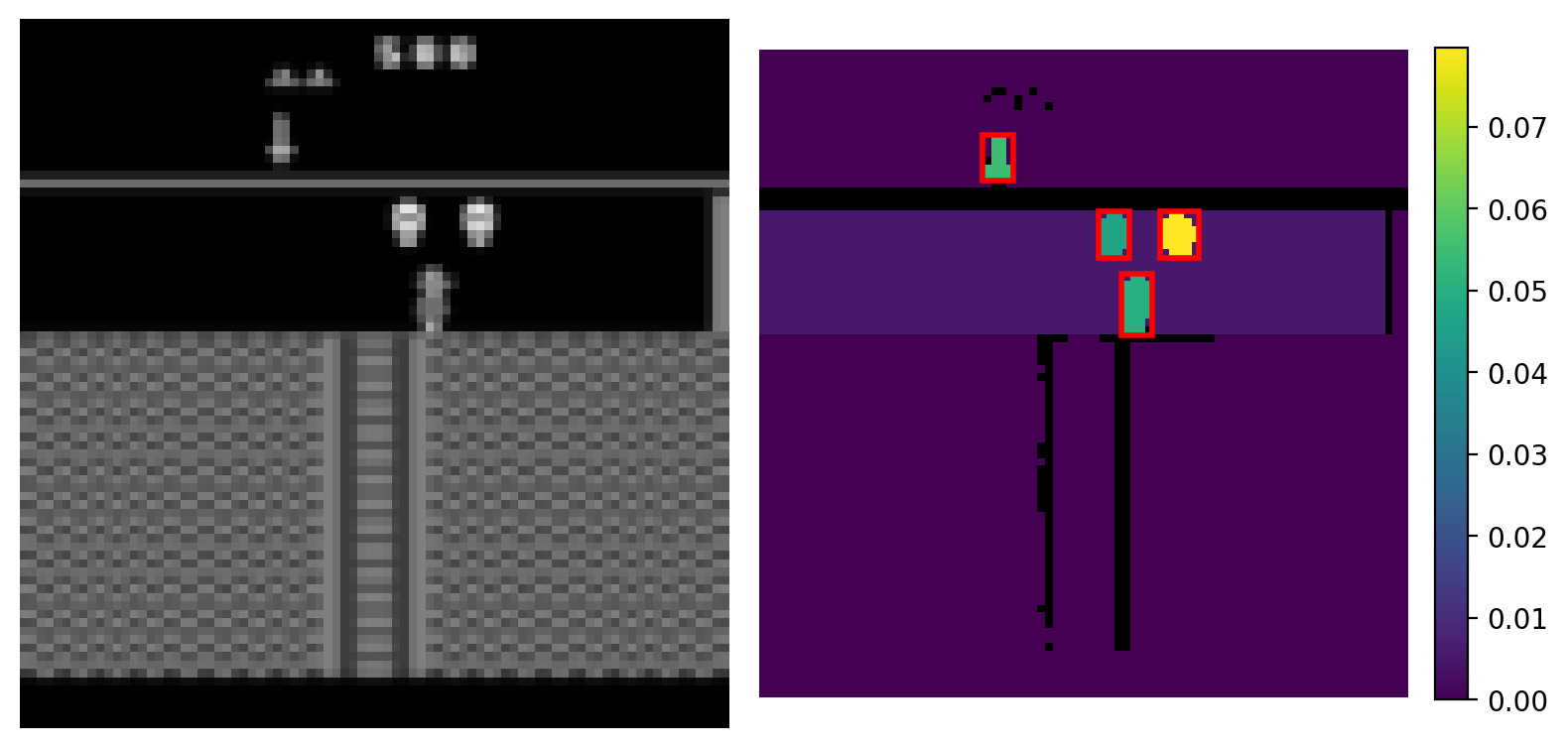}
    \caption{Agent prepares to kill skull with sword.}
    \label{fig:shapley_plots_ext}
\end{subfigure}
\hfill
% \vspace{-0.3cm}
\caption{\textbf{Examples of identifying relevant features in \textsc{MontezumasRevenge}.} For each state (black and white image), we show the estimated Shapley value attributed to each feature in the image and the bounding boxes selected for the subgoal classifier. The first classifier checks if the player has entered room 0, while the second classifier checks if the player has the sword near the skulls.}
\vspace{-0.2cm}
\label{fig:shapley_plots}
\end{figure*}

\paragraph{Feature attribution.} Figure~\ref{fig:shapley_plots} shows two examples of DeepSHAP attributions on discovered subgoals in \textsc{MontezumasRevenge}. The agent enters a new room for the first time, triggering a novelty spike; the intrinsic-model attribution correctly highlights aspects that uniquely identify this room, while ignoring other aspects like the number of lives remaining or the score. Another shows the agent approaching two skulls while equipped with the sword, a state that precedes extrinsic reward. The method isolates meaningful features rather than attending to the entire observation.

\begin{figure*}[h]
    \centering
    \vspace{-0.4cm}
    \includegraphics[width=\linewidth]{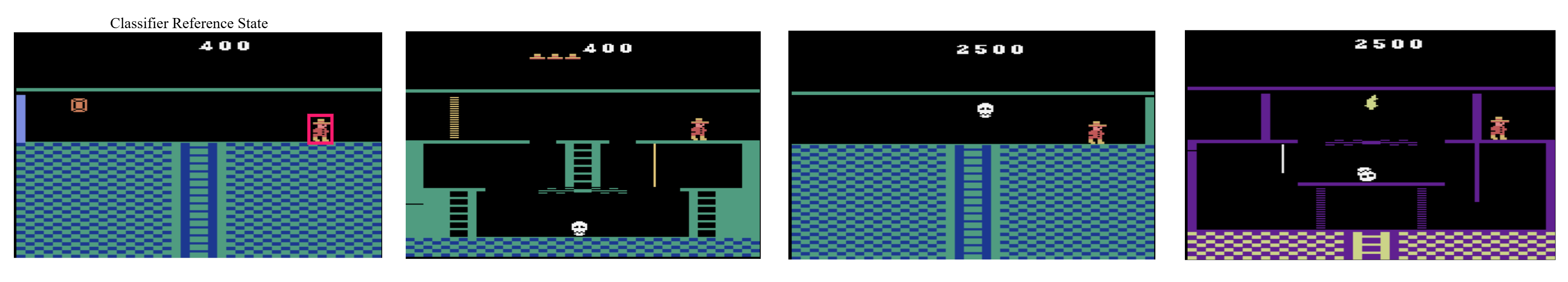}
    \vspace{-0.6cm}
    \caption{\textbf{Example of a discovered subgoal classifier that transfers to many different rooms.} The leftmost image shows the context in which this subgoal classifier was learned---the bounding box shows that the classifier only attends to the player's position. The remaining three images show some states in which that classifier was triggered---as long as the player re-appears in that portion of the screen, other factors are all irrelevant for achieving that option's subgoal.}
    \vspace{-0.5cm}
    \label{fig:example_classifier_monte}
\end{figure*}

\paragraph{Classifier transfer.} Figure~\ref{fig:example_classifier_monte} demonstrates that classifiers built from relevant features transfer across states. The leftmost image shows the state in which a classifier was learned; the bounding box indicates that it attends only to the player's position in a specific screen region. The remaining images show states where this classifier fires later in the agent's lifetime: despite differences in room layout, score, number of lives, and enemy positions, the classifier triggers whenever the player appears in the target region. Appendix~\ref{sec:appendix_minigrid_classifiers} shows examples from \textsc{MiniGrid}, including subgoals that capture the relationship between different objects. Quantitatively, in \textsc{MontezumasRevenge}, we found that a classifier triggered on a median of 71.4 $\pm$ 11 states. Some classifiers are highly selective (e.g., picking up a key while having exactly one life), while others are broadly applicable (e.g., the player appearing in a screen region). 

\subsection{Quantitative Results}
\begin{figure}[h]
    \centering
    \includegraphics[width=\linewidth]{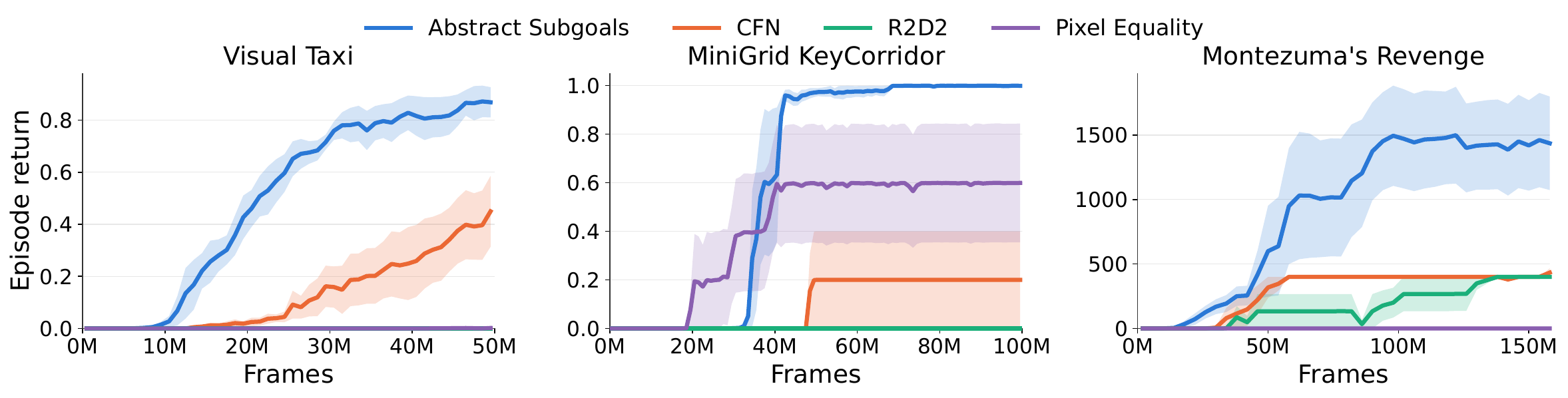}
    \caption{\textbf{Learning curves} comparing our agent (\textcolor{blue}{Abstract Subgoals}) with a non-hierarchical novelty maximizing RL method (\textcolor{orange}{CFN}), a vanilla RL method (\textcolor{green}{R2D2}), and a state-reaching HRL baseline (\textcolor{purple}{Pixel Equality}). Solid lines denote average undiscounted return; shaded regions denote standard deviation over $5$ random seeds.}
    \label{fig:lcs}
\end{figure}

We compare our method (Abstract Subgoals) against three baselines: R2D2 \citep{kapturowski2018recurrent}, a flat RL method with $\epsilon$-greedy exploration; CFN \citep{lobel2023flipping}, a flat RL method with novelty-based exploration; and Pixel Equality, which uses the same online subgoal discovery and training procedure as Abstract Subgoals but replaces the classifier's dissimilarity test $\mathcal{D}$ (Section~\ref{sec:classifier}) with $\sum_{p} (s_N[p] - s[p])^2 < 0.01$, summed over every pixel $p$ of the full image rather than only the selected features $\mathcal{F}$. Figure~\ref{fig:lcs} shows learning curves across all three domains.

Our method, Abstract Subgoals, achieves the best final performance in all three domains (Figure~\ref{fig:lcs}). Among the baselines, CFN outperforms vanilla R2D2, isolating the importance of exploration in these problems. Pixel Equality improves upon CFN in \textsc{MiniGrid-KeyCorridor}, but under-performs in the other two tasks where it is unable to learn transferrable option policies. Our method's improvement over CFN points to the importance of hierarchies, while its improvement over Pixel Equality additionally isolates the importance of feature selection. Appendix~\ref{sec:appendix_pixel_equality_ablation} further isolates the improvement over Pixel Equality.  

\section{Discussion and Conclusion}\label{sec:discussion}
We presented an algorithm that discovers abstract subgoal options by identifying which features of a state account for its novelty, and constructing classifiers that attend only to those features. Our experiments show that this abstraction enables more efficient exploration and better task performance than both flat RL and a state-reaching HRL baseline.

\paragraph{Limitations and future work.} Our method relies on external feature extraction (contour detection, SAM), which is biased toward object-centric features; end-to-end disentangled representation learning \citep{RodriguezSanchez2025FromPT,Kumar2025QuestioningRO} could serve as a drop-in replacement. Not all discovered subgoals are useful: some attend to uncontrollable features like moving enemies, which off-policy controllability estimation \citep{bagaria23scaling} could filter. Our bandit-based policy over options could be replaced with abstract model-based planning \citep{konidaris2018skills}. 

Our key insight is simple---when something interesting happens, figure out \textit{why} it was interesting, not just \textit{that} it was interesting. By grounding option discovery in feature attribution, we take a step toward agents that build abstract, reusable skills from experience.

\bibliography{iclr2027_conference}
\bibliographystyle{apalike}
\newpage
\appendix

\section{Notation}\label{app:notation}

Table~\ref{tab:notation} summarizes the notation used throughout the paper.

\begin{table}[h]
\centering
\caption{Summary of notation.}
\label{tab:notation}
\small
\begin{tabular}{llc}
\toprule
\textbf{Symbol} & \textbf{Meaning} & \textbf{Introduced} \\
\midrule
\multicolumn{3}{l}{\textit{MDP and options}} \\
$M = (S, A, R, T, \gamma)$ & Markov decision process & Sec.~\ref{sec:background} \\
$o = (\mathcal{I}_o, \pi_o, \beta_o)$ & Option: initiation function, policy, termination condition & Sec.~\ref{sec:background} \\
$\mathcal{O}$ & The agent's set of discovered options & Sec.~\ref{sec:background} \\
$H$ & Option timeout (maximum steps per execution) & Sec.~\ref{sec:background} \\
$R_o(s) = \beta_o(s)$ & Option pseudo-reward & Sec.~\ref{sec:background} \\
$V_{\beta_o}, Q_{\beta_o}$ & Option (goal-conditioned) value functions & Sec.~\ref{sec:background} \\
\midrule
\multicolumn{3}{l}{\textit{Exploration and option selection}} \\
$f_\phi^{\text{int}}$ & Learned novelty estimator with parameters $\phi$ & Sec.~\ref{sec:background} \\
$\pi_{\text{int}}, V_{\text{int}}$ & Novelty-driven exploration policy and its value function & Sec.~\ref{sec:exploration} \\
$\pi_\mathcal{O}(o \mid s)$ & Policy over options & Sec.~\ref{sec:exploration} \\
$U(o)$ & Utility of option $o$ under $\pi_\mathcal{O}$ & Eq.~\ref{eq:bandit-utility} \\
$\delta$ & Initiation threshold on $V_{\beta_o}(s)$ & Eq.~\ref{eq:initiation-set} \\
$\alpha, \lambda$ & Extrinsic / intrinsic reward coefficients & Eq.~\ref{eq:bandit-utility} \\
$\tau = (s_1, \ldots, s_K)$ & Exploration trajectory & Sec.~\ref{sec:exploration} \\
$\mu, \sigma$ & Running mean and std.\ of $f_\phi^{\text{int}}$ & Sec.~\ref{sec:exploration} \\
$\sigma_{\text{state}}$ & Subgoal-creation threshold multiplier & Sec.~\ref{sec:exploration} \\
\midrule
\multicolumn{3}{l}{\textit{Feature attribution and subgoal classifiers}} \\
$s_N$ & Most novel state in $\tau$ (single positive example) & Sec.~\ref{sec:exploration} \\
$\mathcal{S}_B = \{s_B^1, \ldots, s_B^M\}$ & Set of $M$ baseline (boring) states & Sec.~\ref{sec:exploration} \\
$w$ & Baseline sampling window around $t_N$ & Sec.~\ref{sec:exploration} \\
$\Delta_n$ & Expected novelty spike of $s_N$ over $\mathcal{S}_B$ & Eq.~\ref{eq:Delta} \\
$\{b_1, \ldots, b_k\}$ & Candidate features extracted from $s_N$ & Sec.~\ref{sec:feature-selection} \\
$c_i$ & Counterfactual state ($b_i$ reset to its value in $s_B$) & Sec.~\ref{sec:feature-selection} \\
$\Delta_i$ & Novelty drop from counterfactual $c_i$ & Sec.~\ref{sec:feature-selection} \\
$\epsilon$ & Counterfactual relevance threshold & Sec.~\ref{sec:feature-selection} \\
$\psi_i$ & Shapley value of feature $i$ & Sec.~\ref{sec:background} \\
$a_p$ & Per-pixel attribution returned by gradient-based methods & Sec.~\ref{sec:feature-selection} \\
$A_i$ & Feature-level attribution (mean of $a_p$ over pixels in $b_i$) & Sec.~\ref{sec:feature-selection} \\
$\mu_A, \sigma_A$ & Mean and std.\ of $\{A_1, \ldots, A_k\}$ & Sec.~\ref{sec:feature-selection} \\
$\sigma_{\text{patch}}$ & DeepSHAP attribution threshold multiplier & Sec.~\ref{sec:feature-selection} \\
$\mathcal{F}$ & Selected subset of relevant features & Sec.~\ref{sec:feature-selection} \\
$g$ & Subgoal classifier (becomes $\beta_o$ of the new option) & Sec.~\ref{sec:classifier} \\
$h$ & Feature extractor used by $g$ & Sec.~\ref{sec:classifier} \\
$\mathcal{D}, \xi$ & Dissimilarity measure and match threshold of $g$ & Sec.~\ref{sec:classifier} \\
$\kappa$ & Classifier trigger window for subgoal filtering & App.~\ref{app:monte} \\
\bottomrule
\end{tabular}
\end{table}

\newpage
\section{Illustration of Core Insight}\label{sec:core-insight}

\begin{figure}[h!!]
    \centering
    \includegraphics[width=0.9\linewidth]{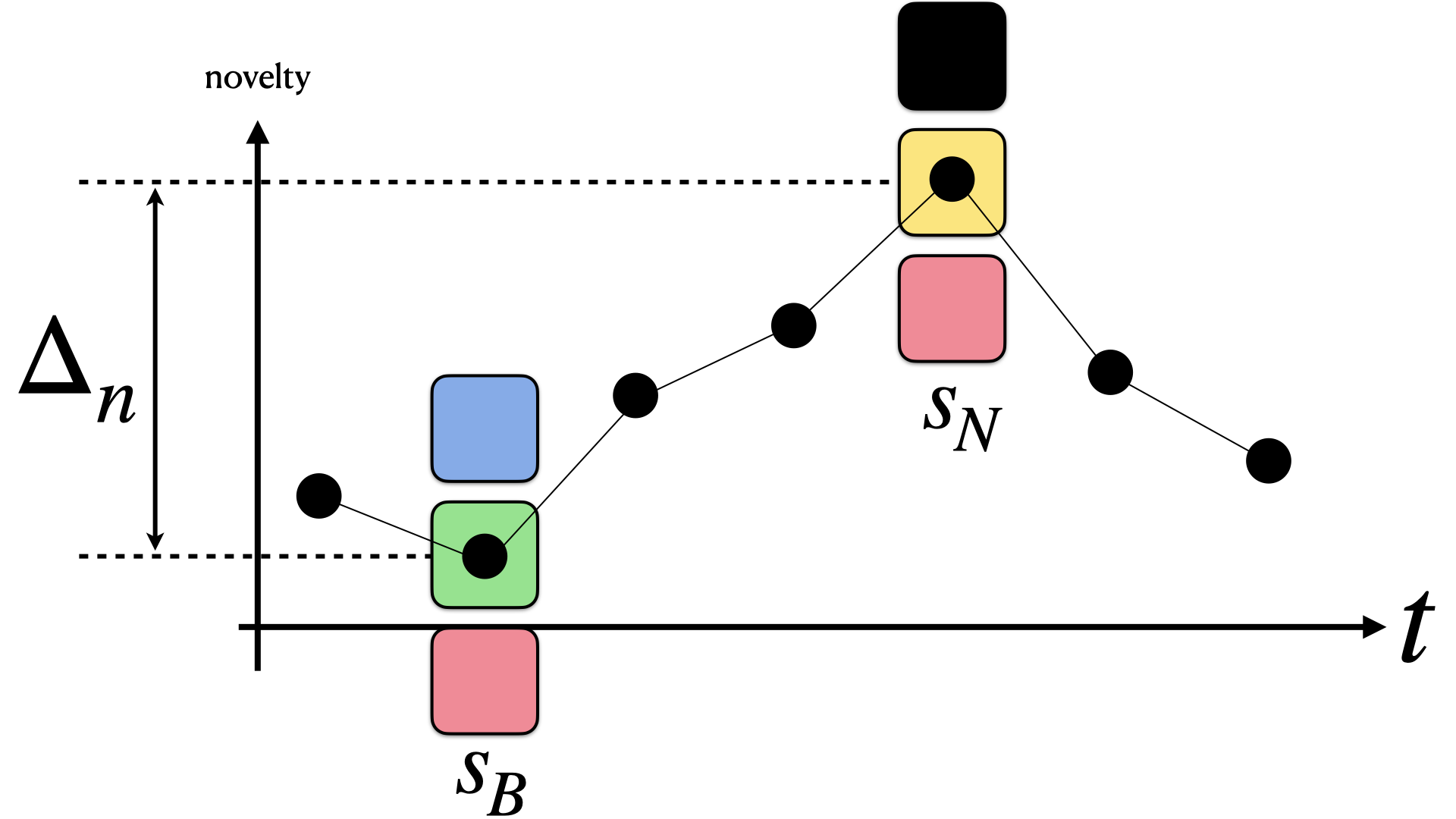}
    \caption[Illustration of abstract subgoal classifiers]{\textbf{Illustration of the core insight.} Suppose an agent observes a state trajectory of length $7$ and a novelty estimator assigns each state $s_t$ in that trajectory a novelty score $f_\phi^{\text{int}}(s_t)$. $s_N$ denotes the most novel state, and $s_B$ denotes the most boring state in that trajectory. Suppose that each state has $3$ features; the colored boxes represent feature values for those two states. Rather than simply treating $s_N$ as a target state, we seek to identify which features in particular were responsible for the large difference in novelty $\Delta_n=f_\phi^{\text{int}}(s_N)-f_\phi^{\text{int}}(s_B)$. Our algorithm can handle input images, but we show state features here for simplicity.}
    \label{fig:factored-clfs/illustration}
\end{figure}

\section{\textsc{MontezumasRevenge} Implementation Details}\label{app:monte}

\paragraph{Initiation function.}  In \textsc{MontezumasRevenge}, instead of using a strict threshold for the initiation function (Equation~\ref{eq:initiation-set}), we instead scale the utility function of the policy over options, $U(o)$ in Equation~\ref{eq:bandit-utility}, by $V_{\beta_o}(s)$. This can be seen as a soft way of modulating the policy over options using the option's initiation function~\citep{HRLSurvey2025}.

\paragraph{Observation normalization.} Following \citet{burda2018exploration}, we whiten CFN's input observations using:
\begin{equation}
    \hat{s}=\text{clip}\left(\frac{s-\mu_{\text{obs}}}{\sigma_{\text{obs}}},-5,5\right)
\end{equation}
where $\mu_\text{obs}$ and $\sigma_\text{obs}$ are the running mean and standard deviation of the observations.
This normalization is applied only to CFN and does not affect the policy or value networks.
This reduces the effect bright objects---such as a white skull---have during pixel attribution, which can bias the contribution of those features.

\paragraph{Window filtering.} In larger, long-horizon domains, our method risks creating too many classifiers.
We add a $\kappa$-window filter: for a novel state $s_N$ encountered at $t_N$ in $\tau$, we discard $s_N$ if any state within $(t_N-\kappa, t_N+\kappa)$ already triggers an existing classifier, avoiding redundant coverage.
This filter's effectiveness depends on classifier generality: a narrowly applicable classifier rarely triggers near other novel states, so few candidate subgoals, if any, are filtered.
It is possible to create a classifier that triggers for a large number of states---for example, a classifier that looks for a specific room or item in the inventory---which would prevent the creation of new subgoals.
We prevent this by excluding classifiers that trigger on every state of the full window duration from consideration during filtering.

\section{Ablations}\label{app:ablations}

\subsection{Hyperparameter Sensitivity}

\begin{figure*}[ht]
\centering
\begin{subfigure}{0.49\textwidth}
    \centering
    \includegraphics[width=\linewidth]{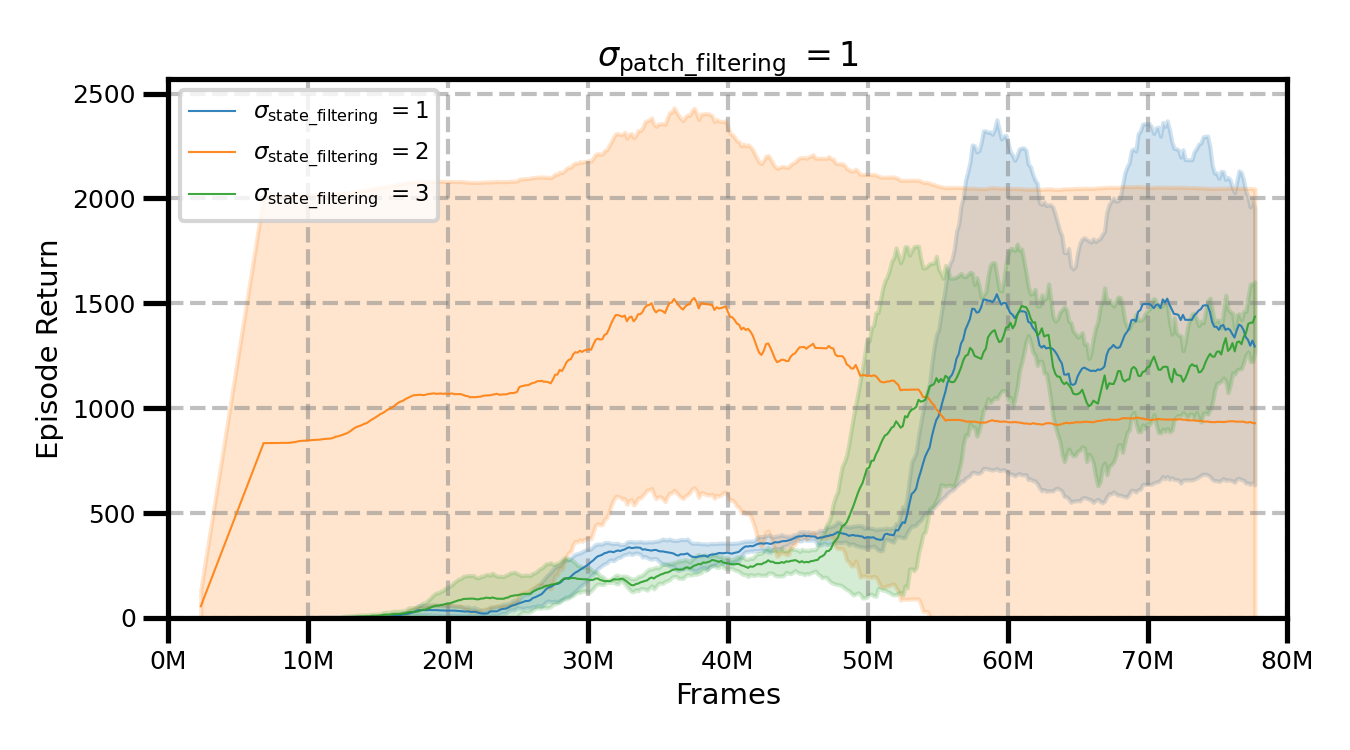}
    \caption{Effect of $\sigma_{\text{state}}$ at fixed $\sigma_{\text{patch}}=1$.}
    \label{fig:ablation_state_sweep}
\end{subfigure}
\hfill
\begin{subfigure}{0.49\textwidth}
    \centering
    \includegraphics[width=\linewidth]{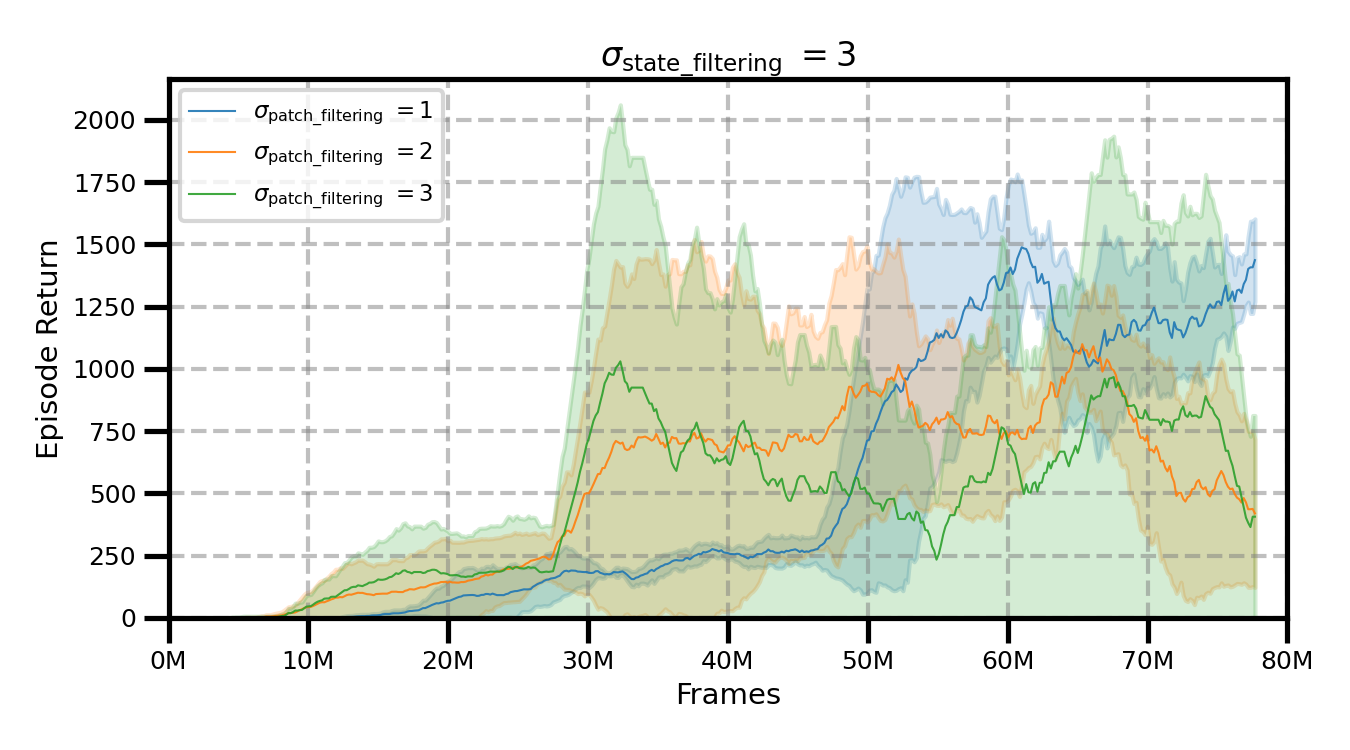}
    \caption{Effect of $\sigma_{\text{patch}}$ at fixed $\sigma_{\text{state}}=3$.}
    \label{fig:ablation_patch_sweep}
\end{subfigure}
\caption{\textbf{Hyperparameter sensitivity sweep in \textsc{MontezumasRevenge}.} Each curve is the average undiscounted task return across 3 random seeds; shaded regions denote standard deviation.}
\label{fig:ablations}
\end{figure*}

We swept the two key threshold multipliers---$\sigma_{\text{state}}$ (the multiplier in the subgoal-creation threshold of Algorithm~\ref{alg:abstract-subgoal-discovery}, for the intrinsic signal) and $\sigma_{\text{patch}}$ (the multiplier in the DeepSHAP attribution threshold of Section~\ref{sec:feature-selection})---each over $\{1\sigma, 2\sigma, 3\sigma\}$ in \textsc{MontezumasRevenge}, with 3 random seeds per configuration. (Our main experiments use $\sigma_{\text{state}}=\sigma_{\text{patch}}=1$.) Figure~\ref{fig:ablations} shows that most configurations reach meaningful scores (1000+), confirming that the method is not overly sensitive to these choices. Lower values of $\sigma_{\text{patch}}$ consistently perform better: more permissive thresholds yield classifiers that generalize more broadly, easing option policy learning. Results for $\sigma_{\text{state}}$ are more mixed, with different values leading at different points during training, suggesting robustness rather than a single optimal setting.

\subsection{Decoupling Discovery from Reachability}\label{sec:appendix_pixel_equality_ablation}

\begin{figure}[h]
    \centering
    \includegraphics[width=0.75\linewidth]{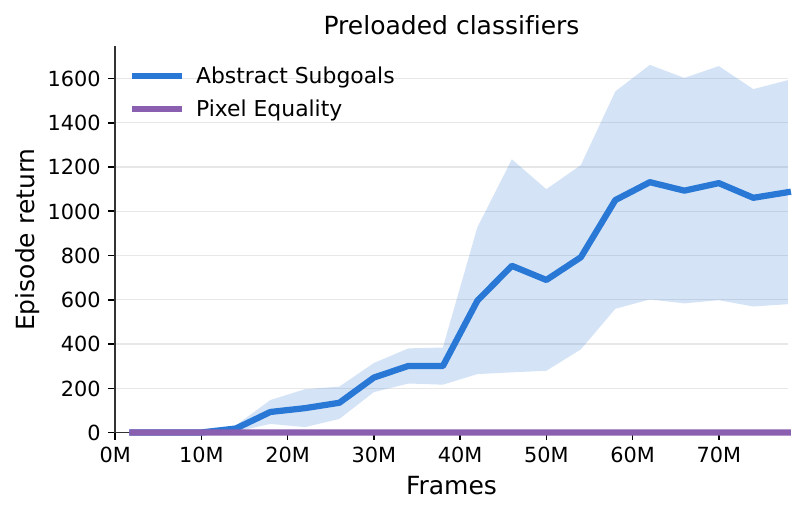}
    \caption{\textbf{Learning curves comparing Abstract Subgoals and Pixel Equality given subgoals from a separate discovery-only stage.} Both agents are trained on the same fixed set of novel trigger states for \textsc{MontezumasRevenge}, with subgoal discovery disabled; each method constructs its own classifier for these states, so the comparison isolates the effect of the classifier alone (averaged over 4 seeds).}
    \label{fig:pixel_equ_comp}
\end{figure}

In Algorithm~\ref{alg:abstract-subgoal-discovery}, the agent only explores for new subgoals after reaching its currently sampled subgoal.
This means a method that reaches its subgoals less reliably also explores less, and so may simply discover fewer or worse subgoals as a result.
The comparisons between Pixel Equality and Abstract Subgoals in Section~\ref{sec:experiments} conflate two distinct effects: (1) how easily each method's classifiers can be reached by the option policy, and (2) which subgoals each method happens to discover.

To isolate effect (1), we run an additional experiment on \textsc{MontezumasRevenge} in which both methods are given an identical, fixed set of subgoals obtained from a separate discovery-only stage: an agent explores purely via RND novelty-driven exploration, with no option selection, producing a fixed set of novel trigger states. Each method then constructs its own classifier for these states using its own procedure---the SAM/DeepSHAP feature-attribution pipeline for Abstract Subgoals, full-image matching for Pixel Equality---so subgoal discovery does not depend on either method's ability to reach previous subgoals.
Subgoal discovery is disabled for the remainder of the experiment, so both methods are trained only to reach the same fixed set of subgoals.
Figure~\ref{fig:pixel_equ_comp} shows that Pixel Equality still achieves an episode return of $0$, while Abstract Subgoals finishes with a final return of $\approx1{,}100$, confirming the advantage of learning abstract subgoals even when both methods are evaluated independently of discovery. The reason for this is likely that the subgoal regions are too small for the non-abstract options to support effective policy learning.

\subsection{Sensitivity to Novelty Estimator}\label{sec:appendix_cfn_vs_rnd}

Our experiments use CFN to estimate novelty but our approach is agnostic to the choice of novelty estimator.
To test whether this choice affects which features get selected, we run an exploration-only discovery stage, using Random Network Distillation (RND;~\citealp{burda2018exploration}) as the novelty estimator and compare the resulting subgoals from RND and CFN.
Since each estimator independently determines its own most-novel states during discovery, RND and CFN are unlikely to agree on which state is most novel at a given point in exploration; we therefore compare the subgoals each estimator creates, rather than forcing a comparison on identical states, which would not reflect each estimator's own judgment of novelty.

\begin{figure}[h]
    \centering
    \includegraphics[width=0.6\linewidth]{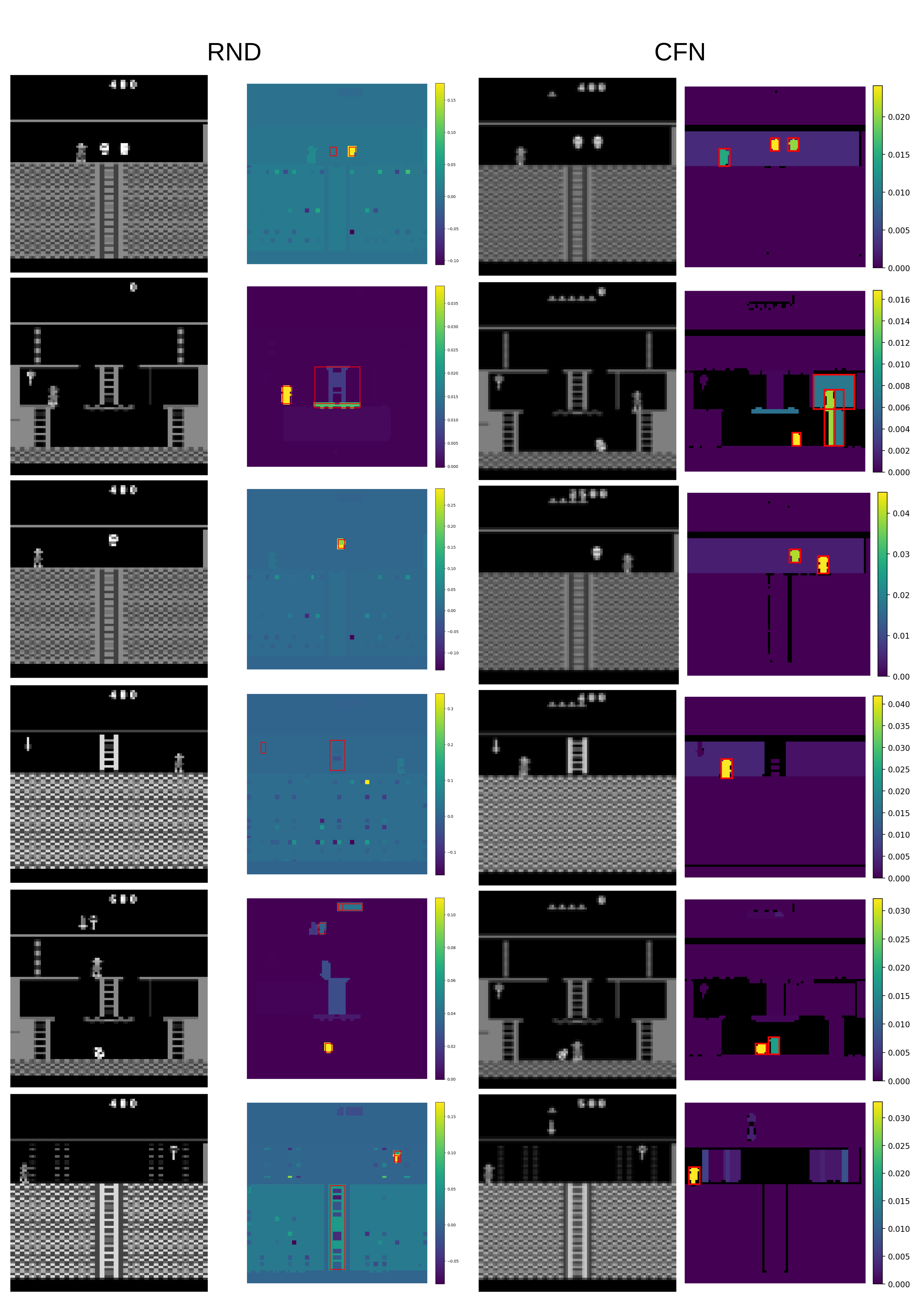}
    \caption{\textbf{DeepSHAP attribution is broadly consistent between CFN and RND novelty estimators, but CFN attribution is more concentrated.} Each row compares the most-novel state independently identified by RND and by CFN at a similar stage of exploration; because the two estimators judge novelty differently, these are not necessarily the same state. The left pair of columns shows RND's most-novel state alongside its DeepSHAP attribution map, and the right pair shows CFN's most-novel state alongside its attribution map. Noticeably, CFN's attributions are more concentrated around a few features, while RND's attributions are more diffused across all features. Despite this, the resulting classifiers attend to similar features.}
    \label{fig:rnd_comp}
\end{figure}

Figure~\ref{fig:rnd_comp} shows that the two novelty estimators produce similar subgoal classifiers: the relevant features largely agree between CFN- and RND-based attribution.
However, the attribution maps differ in how concentrated they are. 
CFN's attribution is close to $0$ over most features for all shown states, with elevated attribution around a few features.
RND shows a similar pattern for the starting room, room $1$; however, for states lying outside this initial room, we see a majority of the features register as moderately novel (light blue), displaying a more diffused attribution map.
Both novelty estimators point to the same relevant features on average, but CFN separates relevant from irrelevant features more cleanly, while RND spreads novelty attribution across more features.

\subsection{Growth in Number of Discovered Options}\label{app:option_growth}

\begin{figure}[h]
    \centering
    \includegraphics[width=0.75\linewidth]{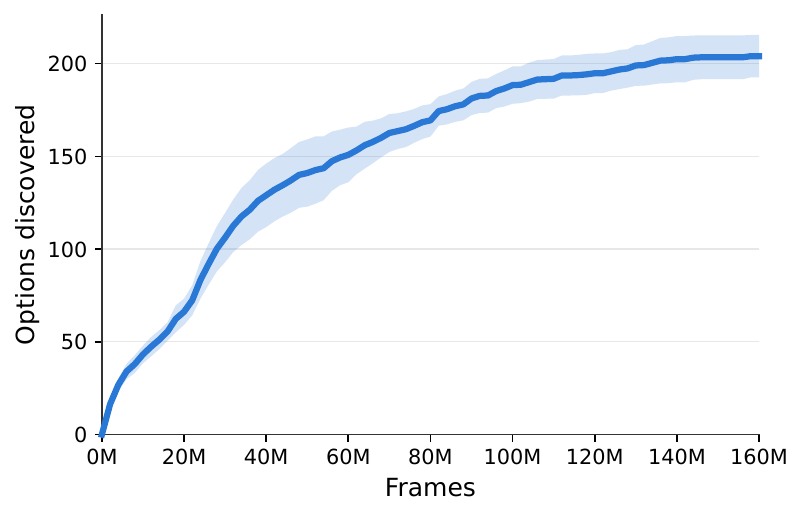}
    \caption{\textbf{Number of options discovered in \textsc{MontezumasRevenge} over training for the learning curves reported in Figure~\ref{fig:lcs}.} Solid line denotes the mean number of options discovered; shaded area denotes standard error.}
    \label{fig:option_growth}
\end{figure}

% One problem motivating this work is that option discovery could continue unchecked, eventually overwhelming the agent with too many options (Section~\ref{introduction}).
Figure~\ref{fig:option_growth} tracks the option count in \textsc{MontezumasRevenge}.
The option count follows a logarithmic trend---rapidly growing initially then continuously decelerating---starting to saturate around $200$ options.
% This shows that classifiers learned early in training continue to be applicable as training progresses, so our window filter (Appendix \ref{app:monte}) prevents creating new, redundant subgoals, guarding against unchecked subgoal growth.

\section{Discovered Subgoals In MiniGrid}\label{sec:appendix_minigrid_classifiers}

\begin{figure*}[h]
    \centering
    \includegraphics[width=.95\linewidth]{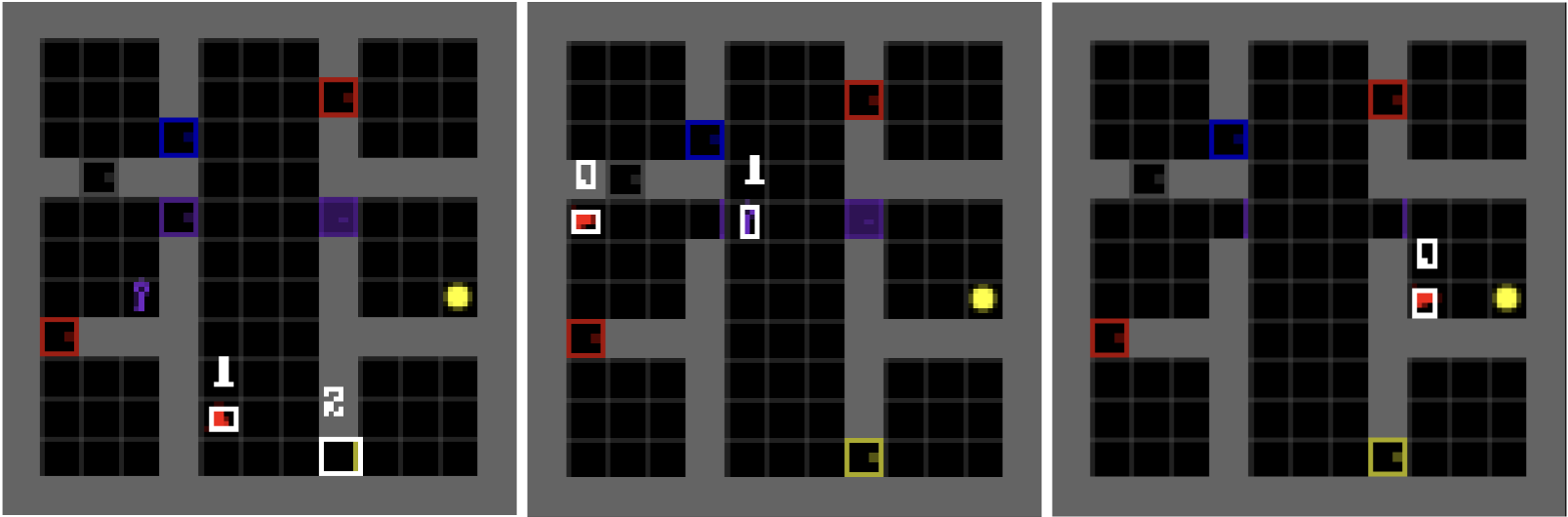}
        \caption{\textbf{Discovered subgoal classifiers in \textsc{MiniGrid-KeyCorridor}.} Bounding boxes show the patches each classifier attends to; all other pixels are ignored. \textit{(Left)} Subgoal: reach a location and open the yellow door. \textit{(Middle)} Subgoal: reach the mid-left room while the key is in the hallway. \textit{(Right)} Subgoal: enter the locked room containing the ball, regardless of key position or door configuration.}
    \label{fig:factored-clfs/kc_clfs}
\end{figure*}

Figure~\ref{fig:factored-clfs/kc_clfs} shows three subgoal classifiers discovered by our agent in \textsc{MiniGrid-KeyCorridor}, each attending only to the patches inside its bounding boxes. Together they illustrate a range of abstraction levels: the left classifier combines the agent's location with the state of the yellow door; the middle one couples the agent's position with the state of the key (which must be in the hallway), capturing a relational subgoal across objects; and the right classifier fires whenever the agent enters the locked room with the ball, regardless of where the key is or whether the door is open. This last example illustrates the kind of generalization that state-reaching subgoals cannot represent---the same subgoal is satisfied across many combinations of irrelevant features.

\clearpage
\section{Hyperparameters}\label{app:hyperparameters}

Hyperparameters for each method were tuned independently, with the exception of Pixel Equality, which intentionally reuses Abstract Subgoals' hyperparameters (Table~\ref{tab:hier-hp}) to isolate the effect of the classifier.

\begin{table}[h]
\centering
\caption{Hyperparameters shared across all agents, domains, and methods, unless overridden in Tables~\ref{tab:flat-baselines}--\ref{tab:option-hp}.}
\label{tab:shared-hp}
\small
\begin{tabular}{ll}
\toprule
\textbf{Hyperparameter} & \textbf{Value} \\
\midrule
\multicolumn{2}{l}{\textit{R2D2 agent}} \\
Number of actors & 32 \\
Actor backend & CPU \\
Sequence batch size & 32 \\
Trace length & 40 \\
Sequence period & 20 \\
Burn-in length & 0 \\
Bootstrap $n$-step & 5 \\
Prefetch size & 1 \\
Samples per insert (default, unless noted) & 2 \\
\midrule
\multicolumn{2}{l}{\textit{Replay buffer}} \\
Max.\ replay size & 100{,}000 \\
Min.\ replay size & 1{,}000 \\
Priority exponent & 0.9 \\
Importance-sampling exponent & 0.6 \\
Max.\ priority weight & 0.9 \\
\midrule
\multicolumn{2}{l}{\textit{CFN module (architecture)}} \\
Output dimension & 20 \\
Batch size & 1{,}024 \\
Max.\ replay size & 2{,}000{,}000 \\
Priority exponent & 1.0 \\
Importance-sampling exponent & 0.6 \\
Max.\ priority weight & 0.9 \\
Predictor learning rate & $1\times10^{-3}$ \\
Variable update period & 400 \\
Extrinsic coefficient & 1.0 \\
\midrule
\multicolumn{2}{l}{\textit{Feature extraction \& attribution (\textsc{KeyCorridor} / \textsc{VisualTaxi})}} \\
Pixel intensity threshold (foreground mask \& contour) & 30 \\
Contour retrieval / approximation mode & \texttt{RETR\_EXTERNAL} / \texttt{CHAIN\_APPROX\_NONE} \\
Box size filter (min.\ width \& height) & $>0.25\times$ tile size (image height / 13) \\
Counterfactual attribution threshold (novelty drop), $\epsilon$ & 0.1 \\
Number of attribution baselines, $M$ & 1 \\
Baseline window, $w$ & episode length \\
Classifier match: mean pixel difference $\leq$ & 60 \\
Classifier match: template score $>$ & 0.5 \\
\bottomrule
\end{tabular}
\end{table}

\begin{table}[h]
\centering
\caption{Flat RL baselines: core hyperparameters per domain. TUP = target update period; SPI = samples-per-insert ratio. ``Reward tx.'' is the transform applied to the R2D2 target (identity or signed-hyperbolic). ``Intrinsic coeff.'' is the coefficient on the CFN novelty bonus added to the training reward.}
\label{tab:flat-baselines}
\resizebox{\textwidth}{!}{%
\begin{tabular}{llccccccccc}
\toprule
Domain & Method & $\gamma$ & LR & TUP & SPI & Reward tx. & Intrinsic coeff. & CFN LR & CFN SPI & CFN min.\ replay \\
\midrule
\multirow{2}{*}{\textsc{Montezuma}} & CFN & 0.99 & $1\times10^{-4}$ & 600 & 2 & signed-hyperbolic & 0.01 & $1\times10^{-3}$ & Unlimited & 2{,}048 \\
 & R2D2 & 0.99 & $1\times10^{-4}$ & 600 & 2 & signed-hyperbolic & 0.0 & --- & --- & --- \\
\midrule
\multirow{2}{*}{\textsc{KeyCorridor}} & CFN & 0.99 & $3\times10^{-4}$ & 1{,}200 & 8 & identity & 0.001 & $1\times10^{-4}$ & Unlimited & 12{,}500 \\
 & R2D2 & 0.99 & $3\times10^{-4}$ & 600 & 2 & identity & --- & --- & --- & --- \\
\midrule
\multirow{2}{*}{\textsc{VisualTaxi}} & CFN & 0.99 & $3\times10^{-4}$ & 600 & 2 & identity & 0.03 & $1\times10^{-4}$ & 8 & 2{,}048 \\
 & R2D2 & 0.99 & $3\times10^{-4}$ & 600 & 2 & identity & --- & --- & --- & --- \\
\bottomrule
\end{tabular}%
}
\end{table}

\begin{table}[h]
\centering
\caption{Abstract Subgoals / Pixel Equality: policy hyperparameters. TUP = target update period; SPI = samples-per-insert ratio. Pixel Equality uses identical values (it differs only on the classifier). $\lambda$ is the CFN exploration-bonus coefficient from Section~\ref{sec:exploration}; it is 0.01 in all three domains.}
\label{tab:hier-hp}
\resizebox{\textwidth}{!}{%
\begin{tabular}{llccccccccc}
\toprule
Domain & Policy & $\gamma$ & LR & TUP & SPI & $\lambda$ & CFN LR & CFN TUP & CFN SPI & CFN min.\ replay \\
\midrule
\multirow{2}{*}{\textsc{Montezuma}} & Intra-option ($\pi_o$) & 0.997 & $1\times10^{-4}$ & 600 & 2 & --- & --- & --- & --- & --- \\
 & Exploration ($\pi_{\text{int}}$) & 0.99 & $3\times10^{-4}$ & 600 & 8 & 0.01 & $1\times10^{-3}$ & 600 & Unlimited & 12{,}500 \\
\midrule
\multirow{2}{*}{\textsc{KeyCorridor}} & Intra-option ($\pi_o$) & 0.997 & $1\times10^{-4}$ & 500 & 2 & --- & --- & --- & --- & --- \\
 & Exploration ($\pi_{\text{int}}$) & 0.99 & $3\times10^{-4}$ & 600 & 8 & 0.01 & $1\times10^{-3}$ & 600 & Unlimited & 12{,}500 \\
\midrule
\multirow{2}{*}{\textsc{VisualTaxi}} & Intra-option ($\pi_o$) & 0.997 & $1\times10^{-4}$ & 600 & 2 & --- & --- & --- & --- & --- \\
 & Exploration ($\pi_{\text{int}}$) & 0.99 & $3\times10^{-4}$ & 600 & 8 & 0.01 & $1\times10^{-3}$ & 600 & Unlimited & 12{,}500 \\
\bottomrule
\end{tabular}%
}
\end{table}

\begin{table}[h]
\centering
\caption{Goal-space and option-discovery hyperparameters for Abstract Subgoals.}
\label{tab:option-hp}
\small
\resizebox{\textwidth}{!}{%
\begin{tabular}{lccc}
\toprule
Hyperparameter & \textsc{Montezuma} & \textsc{KeyCorridor} & \textsc{VisualTaxi} \\
\midrule
Option timeout, $H$ & 100 & 50 & 50 \\
Goal-creation threshold multiplier, $\sigma_{\text{state}}$ & 1 & 1 & 1 \\
Number of goals to replay & 5 & 5 & 5 \\
Option-initiation threshold, $\delta$ & N/A (scaled by $V_{\beta_o}(s)$, Appendix~\ref{app:monte}) & 0.1 & 0.1 \\
Extrinsic-reward weight in $U(o)$, $\alpha$ & 0.25 & 0 & 0 \\
Classifier match threshold, $\xi$ & $0.7$ (with $\mathcal{D} = 1 - \text{IOU}$) & see Table~\ref{tab:shared-hp} & see Table~\ref{tab:shared-hp} \\
\bottomrule
\end{tabular}%
}
\vspace{2pt}

{\small IOU (intersection-over-union) is the ratio between the overlapping area of two bounding boxes and the area of their union.}
\end{table}

\begin{table}[h]
\centering
\caption{\textsc{MontezumasRevenge}-specific classifier discovery hyperparameters.}
\label{tab:monte-hp}
\small
\begin{tabular}{ll}
\toprule
\textbf{Hyperparameter} & \textbf{Value} \\
\midrule
Saliency method & DeepSHAP \\
SAM checkpoint & ViT-H (\texttt{sam\_vit\_h\_4b8939.pth}) \\
Patch attribution threshold, $\sigma_{\text{patch}}$ & 1.0 \\
Number of attribution baselines, $M$ & 15 \\
Baseline window, $w$ & 50 \\
Classifier trigger window size, $\kappa$ & 8 \\
\bottomrule
\end{tabular}
\end{table}

\end{document}